\documentclass[letterpaper, 10 pt, conference]{ieeeconf}  

\IEEEoverridecommandlockouts                              

\usepackage{amsmath}
\usepackage{amsfonts}
\usepackage{graphics} 
\usepackage{graphicx}
\usepackage{todonotes}
\usepackage{array}
\usepackage{multirow}
\usepackage{verbatim} 
\usepackage{mathtools}
\usepackage{amssymb}
\usepackage{url}
\usepackage{tikz}
\usepackage{hyperref}
\usepackage{subfig}
\usepackage{float}

\usepackage{eso-pic}
\newcommand\AtPageUpperMyright[1]{\AtPageUpperLeft{
 \put(\LenToUnit{0.5\paperwidth},\LenToUnit{-1cm}){
     \parbox{0.5\textwidth}{\raggedleft\fontsize{9}{11}\selectfont #1}}
 }}
\newcommand{\conf}[1]{
\AddToShipoutPictureBG*{
\AtPageUpperMyright{#1}
}
}

\conf{This Article has been accepted in  IEEE International Conference on Multisensor Fusion
and Integration (MFI 2024),  4 – 6 September 2024, Pilsen, Czechia. Kindly cite accordingly} 

\title{\LARGE \bf
Efficient Frontier Management for Collaborative Active SLAM
}

\author{ Muhammad Farhan Ahmed$^{1}$, Matteo Maragliano$^{2}$, Vincent Frémont$^{1}$, Carmine Tommaso Recchiuto$^{2}$, \\Antonio Sgorbissa$^{2}$
\thanks{*This work was carried out in the framework of the NExT Senior Talent Chair DeepCoSLAM, which was funded by the French Government, through the program Investments for the Future managed by the National Agency for Research ANR-16-IDEX-0007, and with the support of Région Pays de la Loire and Nantes Métropole.}
\thanks{$^{1}$ Muhammad Farhan Ahmed, Vincent Frémont are with École Centrale de Nantes, LS2N, Nantes France}%
\thanks{$^{2}$ Matteo Maragliano, Carmine Tommaso Recchiuto, Antonio Sgorbissa are with University of Genoa, DIBRIS Department, Genoa Italy}%
}

\begin{document}
\maketitle
\thispagestyle{empty}
\pagestyle{empty}

\begin{abstract}
In autonomous robotics, a critical challenge lies in developing robust solutions for Active Collaborative SLAM, wherein multiple robots collaboratively explore and map an unknown environment while intelligently coordinating their movements and sensor data acquisitions. In this article, we present an efficient centralized frontier sharing approach that maximizes exploration by taking into account information gain in the merged map, distance, and reward computation among frontier candidates and encourages the spread of agents into the environment. Eventually, our method efficiently spreads the robots for maximum exploration while keeping SLAM uncertainty low. Additionally, we also present two coordination approaches, synchronous and asynchronous to prioritize robot goal assignments by the central server. The proposed method is implemented in ROS and evaluated through simulation and experiments on publicly available datasets and similar methods, rendering promising results.
\end{abstract}

\section{INTRODUCTION}
\label{sc: introduction}
Autonomous robotics has emerged as a transformative force in the exploration of complex and uncharted environments. From planetary exploration missions to disaster relief operations, the deployment of autonomous robots has demonstrated a revolutionary potential across a diverse range of applications. At the heart of this success lies the robot's ability to autonomously explore an environment while gathering data and constructing detailed maps of the surrounding environment in real-time—a process known as Active Simultaneous Localization and Mapping (A-SLAM). 

Many research works have recently focused on Active Collaborative SLAM (AC-SLAM), which capitalizes on the power of multiple robots working in collaboration. The potential advantages are manifold, from accelerated mapping of terrains to resilient operation in challenging and dynamic scenarios. However, the utilization of multiple robots in collaborative SLAM is not without its challenges. Coordination, resource allocation, and sensor fusion become critical facets that demand careful consideration. Furthermore, the seamless integration of individual robot efforts into a coherent, unified map poses a non-trivial computational and algorithmic challenge. 

We propose an implementation of an AC-SLAM algorithm and extend the work in \cite{placed_fast_2021} to a multi-agent system, where multiple robots collaboratively map an environment. To achieve this aim, we propose an effective method to distribute robots in the environment hence favoring exploration and considering agent priorities using reward, distance-based, and merged map information gain metrics to optimize goal selection. We also propose two communication strategies namely synchronous and asynchronous respectively in a centralized approach with a central server, to establish effective communication and coordination of goals among the agents. We implement the proposed approach in ROS using the client-server model and provide extensive simulation results and real world experiments. 

The subsequent sections are organized as follows: Section \ref{sc: related_work} provides a review of related work, Section \ref{sc: methodology} explains the methodology of the proposed approach, Section \ref{RESULTS} shows the simulation and experimental results, finally we conclude Section \ref{sc: conclusions} summarizing our contributions and prospects for future work.

\section{RELATED WORK}
\label{sc: related_work}
\subsection{Active SLAM}
In A-SLAM, the robot can actively choose its actions, such as selecting views or locations to investigate, to reduce the uncertainty of its localization and map representation for environment exploration. Thus intelligently planning and executing robot operations to minimize uncertainty, with the objective to increase the efficiency and accuracy of SLAM as described in \cite{carrillo_comparison_2012}, \cite{feder_adaptive_1999}. 
 Once the robot has established a map of its surroundings, it proceeds to locate frontier points.  \cite{yamauchi_frontier-based_1997} defines frontier as the boundary separating known map locations from unknown ones, as observed by the robot's sensors. After identifying these goal frontiers, the robot computes a cost or utility function. This function relies on the potential reward associated (optimality criterion) as debated in \cite{placed_explorb-slam_2022} with selecting the optimal action from a set of all possible actions. Theory of Optimal Experimental Design (TOED) \cite{TOED} and concepts from Information Theory (IT) \cite{it} are used to provide optimality criterion for reward computation by the utility function. TOED is used to provide a scalar mapping of pose graph covariance matrix as described in  \cite{carrillo_autonomous_2015} and \cite{carrillo_comparison_2012} debating on its determinant and Eigenvalues (D-Optimally criterion) to guide the reward function to the goal location. While in IT joint Entropy is used. Interested readers are guided to \cite{s23198097}, \cite{placed_general_2023}, \cite{khosoussi_reliable_2019}, \cite{khosoussi_novel_2014} for discussion of uncertainty quantification methods.

\subsection{Frontiers-based Approaches}
 Frontiers play a pivotal role in augmenting the precision of robot localization by enabling intelligent exploration and data acquisition strategies, effectively reducing uncertainty, and enhancing the map-building and localization processes. In \cite{BATINOVIC20209682} an active exploration strategy is proposed where each frontier is weighted based on distance and surrounding unknown cells. While in \cite{7125079} each frontier is segmented, a trajectory is planned for each segment, and the trajectory with the highest map-segment covariance is selected from the global-cost map. The work presented in \cite{dariothesis} uses frontier exploration for autonomous exploration a utility function based on Shannon's and Renyi entropy is used for the computation of the utility of paths. The method described by \cite{CS14} uses a cost function that is somewhat similar to \cite{AC15}, which takes into consideration the discovery of the target area of a robot by another member of the swarm and switches from a frontier to a distance-based navigation function to guide the robot toward the goal frontier.
  Frontiers-based coverage approaches in \cite{AC8} divide the perception task into a broad exploration layer and a detailed mapping layer, making use of heterogeneous robots to carry out the two tasks while solving a Fixed Start Open Traveling Salesman Problem (FSOTSP). Once a frontier has been identified, the robot can use path planning algorithms to reach it and maximize the exploration while minimizing its SLAM uncertainly.

\subsection{Active Collaborative SLAM}

In AC-SLAM, the frontier detection and uncertainty quantification approaches described earlier are also applicable with additional constraints of managing computational and communication resources, and the ability to recover from network failure. The exchanged  parameters are entropy \cite{dariothesis} \cite{AC17}, Kullback–Leibler Divergence (KLD) \cite{kld,AC18}, localization info \cite{AC15}, visual features \cite{AC16}, and frontier points. The authors of \cite{AC2, AC30}, incorporate these multirobot constraints by adding the future robot paths while minimizing the optimal control function which takes into account the future steps and observations and minimizing the robot state and map uncertainty and adding them into the belief space (assumed to be Gaussian).

 \cite{AC17} presents a decentralized method for a long-planning horizon of actions for exploration and maintains estimation uncertainties at a certain threshold. The active path planner uses a modified version of RRT* and an action is chosen that best minimizes the entropy change per distance traveled. The main advantage of this approach is that it maintains good pose estimation and encourages loop-closure trajectories. An interesting solution is given by a similar approach to the method proposed by \cite{AC18} using a relative entropy (RE)-optimization method which integrates motion planning with robot localization and selects trajectories that minimize the localization error and associated uncertainty bound. A planning-cost function is computed, which includes the uncertainty in the state in addition to the state and control cost.

When considering multi-robot systems, two primary aspects come into play. Firstly, teams can be either homogeneous, consisting of robots of the same type, or heterogeneous, \cite{lajoie_towards_2022}, with various robot types working together. Secondly, the system's architecture can be centralized, decentralized, or distributed, \cite{s23198097}. Centralized control offers precise coordination but is susceptible to delays and single points of failure. On the other hand, decentralized systems distribute control for enhanced robustness and scalability while requiring effective coordination. Distributed systems empower individual robots for autonomous decision-making, providing fault tolerance and adaptability while demanding efficient communication protocols. Sometimes, systems can combine centralized and distributed elements \cite{lajoie_towards_2022}, sharing computational tasks among agents while central nodes handle decision-making.

 \section{METHODOLOGY}
\label{sc: methodology}
 While many research works have been focused on collaborative strategies for SLAM, or single-robot active-SLAM, only a few works have dealt with AC-SLAM. However, these approaches present common limitations: a) they have high computational costs associated with the number of frontiers processed. b) They fail to encourage the spread of robots into the environment. c) The uncertainty is quantified by a scalar mapping of the entire pose graph covariance matrix which may become very large, especially in landmark-based SLAM methods increasing the computational cost.  Furthermore, they do not explicitly implement strategies for efficient management of frontiers to speed up map discovery and robot localization. In this work, we propose an AC-SLAM approach that deals with overcoming these limitations. Our proposed method outlines a strategy aimed at reducing in number of frontiers for reward computation and distributing robots within the environment, thereby facilitating exploration and mapping. We leverage a combination of reward metrics, distance-based evaluations, and merged map information gain to refine the goal selection.

In the context of single robot A-SLAM the work of \cite{placed_fast_2021} uses the Open Karto\footnote[1]{\url{http://wiki.ros.org/open_karto}} in ROS noetic\footnote[2]{\url{http://wiki.ros.org/noetic} .} as SLAM Back-end and proposes a modern D-Optimality criterion which is not computationally expensive for uncertainty quantification. This D-optimality criterion is computed as the maximum number of spanning trees in the weighted graph Laplacian. The reward for each frontier candidate is weighted by this optimality criterion and is passed to the path planner to guide the robot to perform A-SLAM. For a set of frontiers $ F = \{f_0, f_1,\ldots,f_N\} \subset \mathbb{R}^2, \text{ where } \forall i \in {0,1, \dots, N}, f_i = (x_i,y_i) $, each robot computes a matrix of rewards $ R = \{r_0, r_1,.....r_N\} \in \mathbb{R}$ as shown in Equation \ref{eq: reward_matrix}. 

\begin{equation}
\label{eq: reward_matrix}
R = 
    \left[
        \begin{array}{ccc}
             \text{Reward} & \text{X} & \text{Y} \\
            \hline
            r_0 & x_0 & y_0 \\
             \vdots & \vdots & \vdots \\
            r_N & x_N & y_N \\
        \end{array}
    \right]
\end{equation}

In this article, we expand \cite{placed_fast_2021} to a multi-robot AC-SLAM and propose an efficient frontier sharing and exploration method. We propose two exploration approaches namely Synchronous and Asynchronous respectively \ref{subsec: synch_asynch_app} for goal assignment to robots. Additionally, we also present an efficient spread policy \ref{subsec: spread_policy} for encouraging exploration.

\begin{figure}
    \centering
    \includegraphics[height=3cm ,width=7cm]{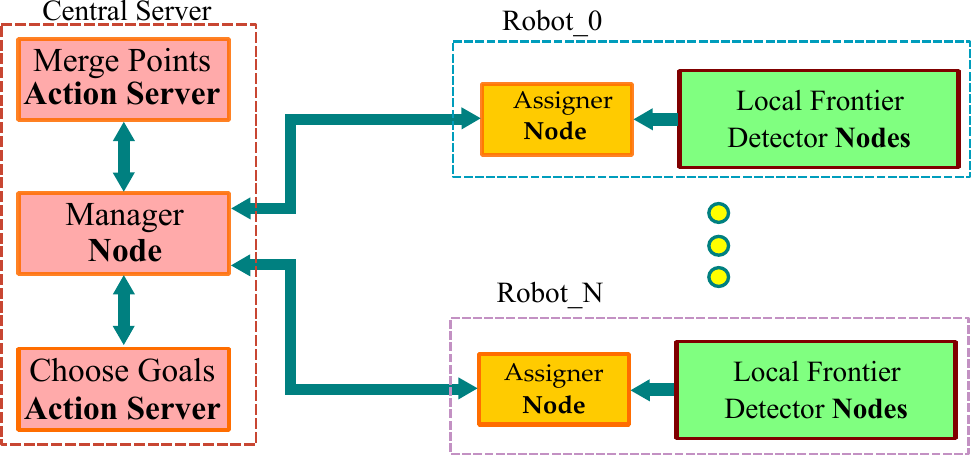}
    \caption{Central server (red), and local nodes (yellow, green) communication.}
    \label{fig: nodes_server}
\end{figure}

\begin{figure}
    \centering
    \includegraphics[height=5.2cm ,width=8.9cm]{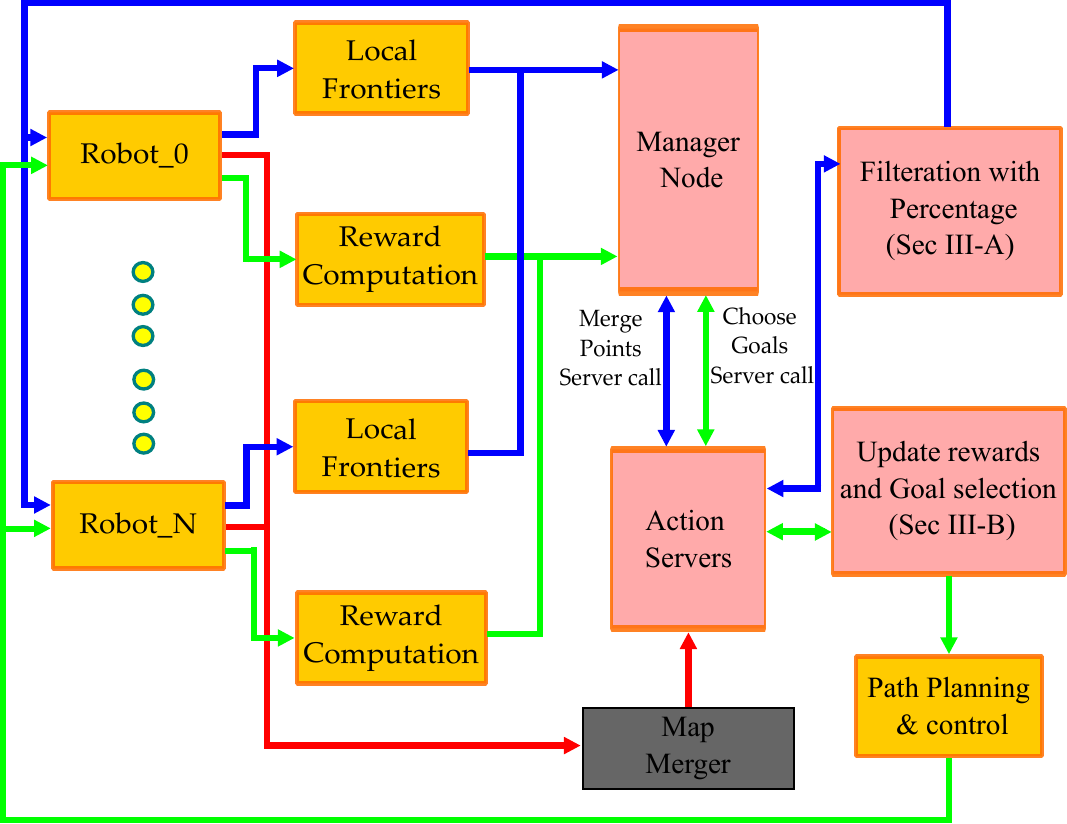}
    \caption{The architecture of resultant system.}
    \label{fig: architecture}
\end{figure}

We developed our proposed approach in ROS Noetic using the ROS actionlib\footnote[3]{\url{http://wiki.ros.org/actionlib}} library. We add a \textit{central server} that receives the list of local frontier points from each robot, computes a global list, and replies with the next target to be reached by the robot. As shown in Figure \ref{fig: nodes_server} each robot detects local frontiers in its map using OpenCV and RRT-based frontier detection from \cite{placed_fast_2021} and passes them to the \textit{manager node} which acts as a communication gateway between the server and robot. The \textit{merge points} action server creates a unique list of frontier points to be used by all the agents and \textit{choose goals} action server chooses the best goal position for each agent depending on the reward matrix as shown in Equation \ref{eq: reward_matrix} and spread criterion. Finally, the \textit{assigner} node receives the chosen goal frontier and executes the path planning action using Dijkstra’s algorithm and DWA planners from the ROS navigation stack\footnote[4]{\url{http://wiki.ros.org/navigation}} as local and global planners respectively. Figure \ref{fig: architecture} shows the resultant architecture of the proposed method in ROS with their namespaces. The orange and pink nodes represent the \textit{assigner node} and central server nodes from Figure \ref{fig: nodes_server}. The grey node is map merging responsible for taking the local maps from each agent and computing a merged map\footnote[5]{\url{http://wiki.ros.org/multirobot_map_merge}}. Filtering with percentage and update rewards \& goal selection will be explained in Sections \ref{subsec: frontiers_management} and \ref{subsec: spread_policy} respectively. Throughout this article, we will use the words \textit{robots} or \textit{agents} interchangeably, and the same applies to \textit{frontiers} and \textit{points}, as they imply the same meaning in the context.
In the following Sections, we elaborate the management policy of the frontiers (Section \ref{subsec: frontiers_management}) and the spreading policy used to speed up the exploration (Section \ref{subsec: spread_policy}) are discussed. We also present two communication methodologies i.e., synchronous and asynchronous (Section \ref{subsec: synch_asynch_app}) which deal with the goal assignment to robots.   

\subsection{Frontiers Management}
\label{subsec: frontiers_management}
\begin{figure}
    \centering
    \includegraphics[height=2.5cm ,width=6cm]{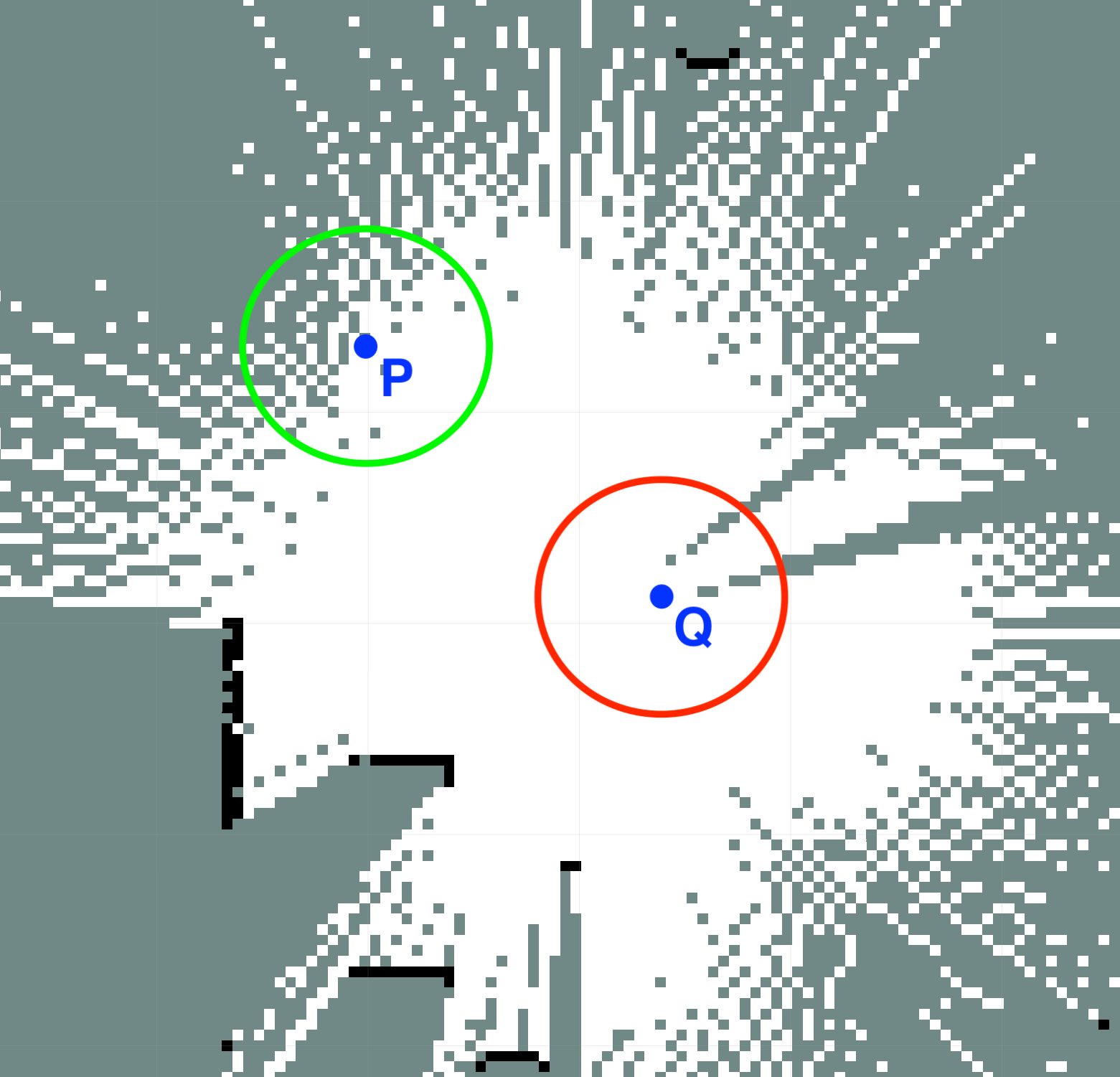}
    \caption{O.G map representing two points and their radius.}
    \label{fig: mapped_area_circle}
\end{figure}

Each agent identifies a certain list of frontier points that will be merged on the server side. Depending on the extent of the map, the final global list may consist of several points, which can lead to high computational time on the server side. For this reason, a strategy to reduce the overall number of frontiers was developed. Also, since we are working on multiple robots, some of the points that are considered frontiers in a local map will be located in a region that is fully mapped when considering the global map. To solve both the aforementioned problems, we decided to consider only those points that have a given percentage of unknown cells within a given radius, using a discretized circle and the global merged map.
 For each detected frontier $f \in F$ in the robot frame $R_f$ we compute the homogeneous coordinates in the merged map as $f_{m} = T_{m}^{Rf}f_{Rf}$ where $f_{Rf} \in \mathbb{R}^4$ is the homogeneous coordinates in robot frame. $T_{m}^{Rf} \in \mathbb{R}^{4\times4}$ is the transformation matrix between merged map and robot frame.     
Figure \ref{fig: mapped_area_circle} shows an example list of two points \textit{P} and \textit{Q} in a partially discovered global Occupancy Grid (O.G) map. For both points, a circle of known radius \texttt{RAD} is drawn and we compute the percentage of the unknown cells over the total inside the circle. Once the percentage is computed, this point is kept or discarded based on the threshold of \texttt{PER\_UNK} set. In the specific case, opportunely setting \texttt{PER\_UNK}, point \textit{P} will be added to the global list, i.e., considered as a border point, whereas point \textit{Q} will be discarded.

The usage of a discretized circle can lead to an Inclusion error: the discretized circle may include some cells outside the circular boundary. This error leads to false positives.
Exclusion error: the discretized circle may exclude some cells within the actual circular boundary. This error leads to false negatives.

The magnitude of the error depends on the resolution used for the O.G map: higher resolutions provide a more accurate approximation of the circle and consequently negligible errors. Unfortunately, using this approach to reduce the list of frontier points (points) may not be sufficient to meet time constraints on the server side. Therefore, we devised an algorithm aimed at further reducing the point number through the adjustment of the radius considered before. The algorithm checks if the global number of points in the list is above a certain threshold \texttt{MAX\_PTS}: in this case, it recomputes a new list of frontiers by increasing the radius \texttt{RAD} to 0.25$m$; conversely, if the number of points is below a fixed threshold \texttt{MIN\_PTS}, the list is reprocessed by decreasing \texttt{PER\_UNK} by 10\% which is the a-priori fixed given percentage of unknown cells on a given radius. This strategy gives a sufficient number of frontier points in the list for the robots to compute their reward. 

\subsection{Spread Policy}
\label{subsec: spread_policy}
To choose navigation targets that allow the agents to explore the map efficiently, a specific spread policy has been implemented. The server keeps track of the already assigned goals. When a target goal for one agent is selected, it updates the reward $R_{new}$ for all other agents from the old reward $R_{old}$, by using a subtracting factor $k$ as shown in Equation \ref{eq: R_update}.

\begin{minipage}{.5\linewidth}
  \begin{equation}
\label{eq: R_update}
    R_{new} = R_{old} - k
  \end{equation}
\end{minipage}%
\begin{minipage}{.3\linewidth}
  \begin{equation} \label{eq:j}  
    k = \frac{K}{d^2}
  \end{equation}
\end{minipage}

\begin{equation}
\label{eq: K}
    K = \frac{\texttt{max reward}}{\texttt{number of targets assigned}}
\end{equation}

The numerator $K$, in Equation \ref{eq:j}, is set at run-time since it depends on the maximum reward for each agent, and the number of targets already assigned. The denominator $d$ represents the Euclidean distance computed between the last chosen goal and the frontier points in the matrix. In other words, when the server assigns a target to robot $j$, it will reprocess all the reward matrices for the other agents, updating the reward with a subtractive factor $k$, which strictly depends on the position of the target assigned to robot $j$. 

Since the $k$ is inversely dependent on $d$, the closer the points are to already chosen goals the less likely they are to be chosen as the next goals, thus achieving the task of spreading the goals in the environment. Normalizing $K$ in Equation \ref{eq: K} with the size of the rewards in each matrix, allows for having a subtractive factor that is scaled with respect to the reward matrix of each agent. Thus taking into account the number of already selected points, possibly distributing the reward "budget" among them. By dividing the maximum reward by the total number of selected points, when the number of targets already explored becomes significant each point will only receive a smaller portion of the total reward, resulting in a more limited effect of $k$.
In the case of asynchronous approach discussed in Section \ref{subsec: synch_asynch_app}, the priority assigned to robots can lead to having one or more robots with low priority being stuck because they are always prioritized by higher-priority agents. To avoid this issue, the server also takes into account the number of requests not related to each agent. Once this number exceeds a certain predefined threshold \texttt{GOAL\_SKIP\_WAIT}, the corresponding agent will be associated with the higher priority. This approach will avoid having robots stuck, distributing goals more uniformly.

\subsection{Sychronous and Asynchronous Approach}
\label{subsec: synch_asynch_app}
The communication between the agents and the server has been implemented with two policies: synchronous and asynchronous. In the synchronous approach, during the execution of the program, each agent receives the same number of goals. Moreover, each agent waits for all the other robots in the system to reach their goal before starting a new goal procedure. In this case, the central server (Figure \ref{fig: nodes_server}) has to manage $n$ different agents at the same time and, during the reward computation, the server is given $n$ Reward Matrices, (Equation \ref{eq: reward_matrix}) one for each robot. A priority among agents has been set so that goal assignment is performed respecting this sequence: given two agents $i$ and $j$ with $i < j $, $i$ is assigned a goal before $j$. 

In the asynchronous approach, each agent is assigned in sequence as many goals as it can reach, without waiting for other agents. In this case, the priority is used to choose the winning agent in the case multiple agents perform the request at the same time. Since with this policy, an agent with a low priority can be stuck for a long time, there is also a counter that keeps track of this prioritization among the robots and, when an agent with a low priority is not considered for a long time, automatically assigns it the highest priority so as that the server will satisfy its request as soon as possible. Since the server is used once at a time and the goal chosen is for one robot at a time, the server for each goal of the agent $i$ stores it; this allows the server not to choose an already chosen goal and to update the rewards to spread the agents taking into account all the goals set so far.
 
\section{RESULTS}
\label{RESULTS}
\subsection{Simulation Results}
\label{sc: simulation_results}
\begin{figure}
    \centering
      \subfloat[W.G\label{fig: plugin_map:1a}]{%
           \includegraphics[height=3cm ,width=4.1cm]{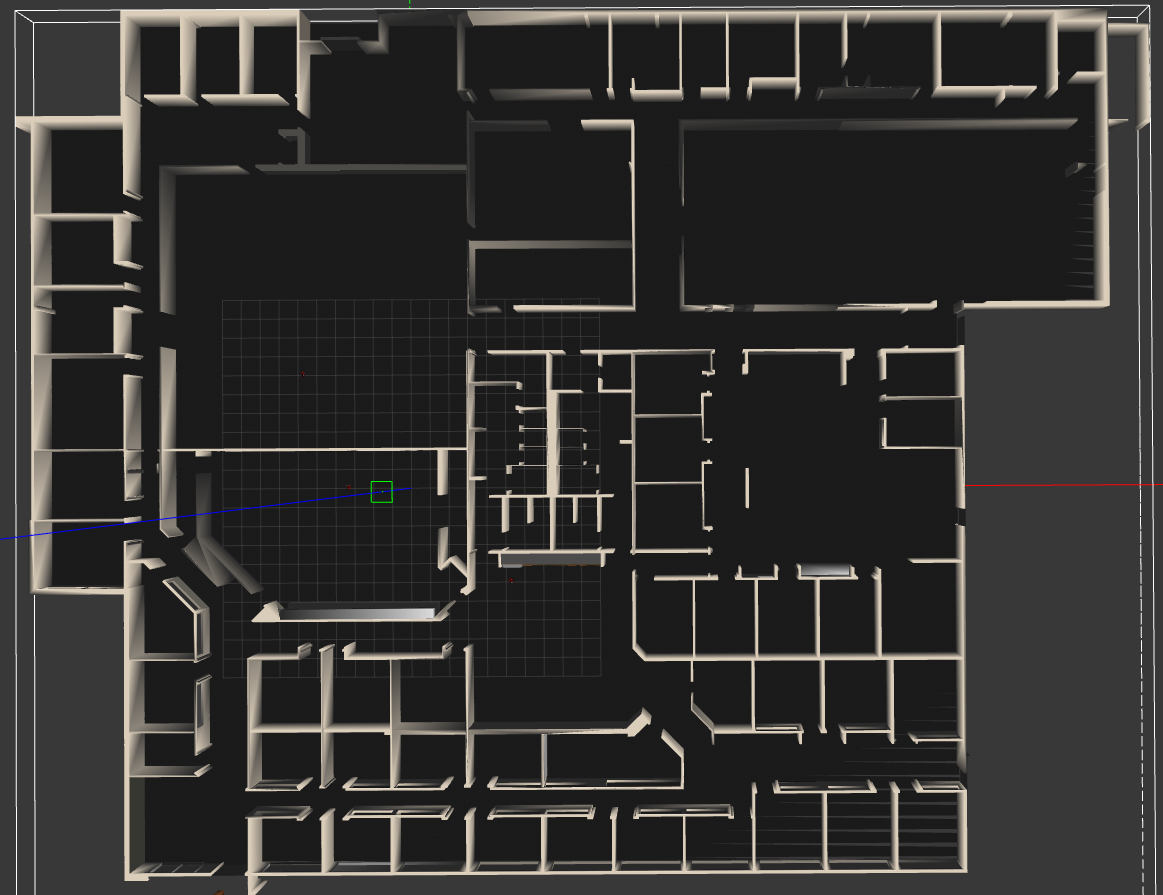}}
        \hfill
      \subfloat[HOS \label{fig: plugin_map:1b}]{%
            \includegraphics[height=3cm ,width=4.4cm]{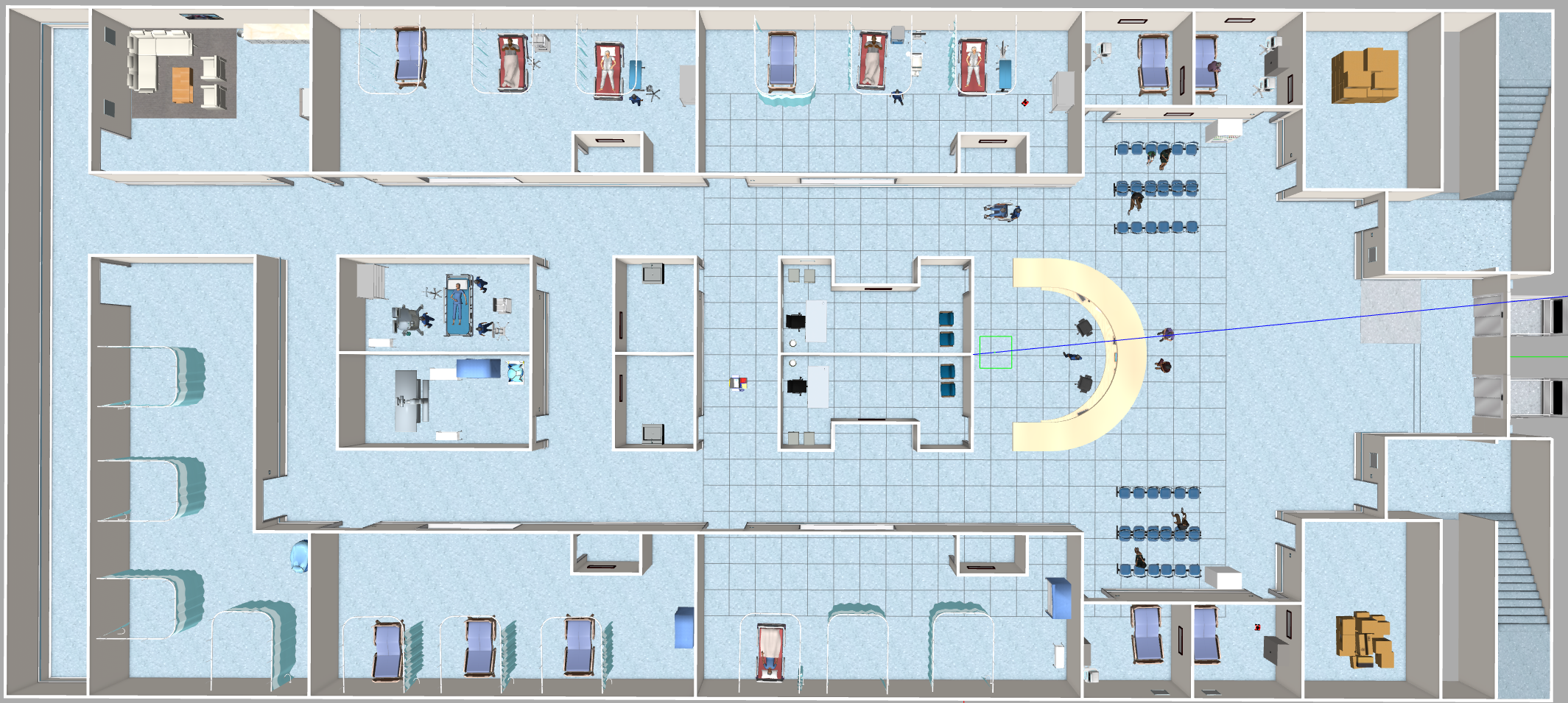}}   

      \subfloat[Resulting O.G map of AWS Hospital environment.\label{fig: plugin_map:1c}]{%
           \includegraphics[height=4.7cm ,width=8.6cm]{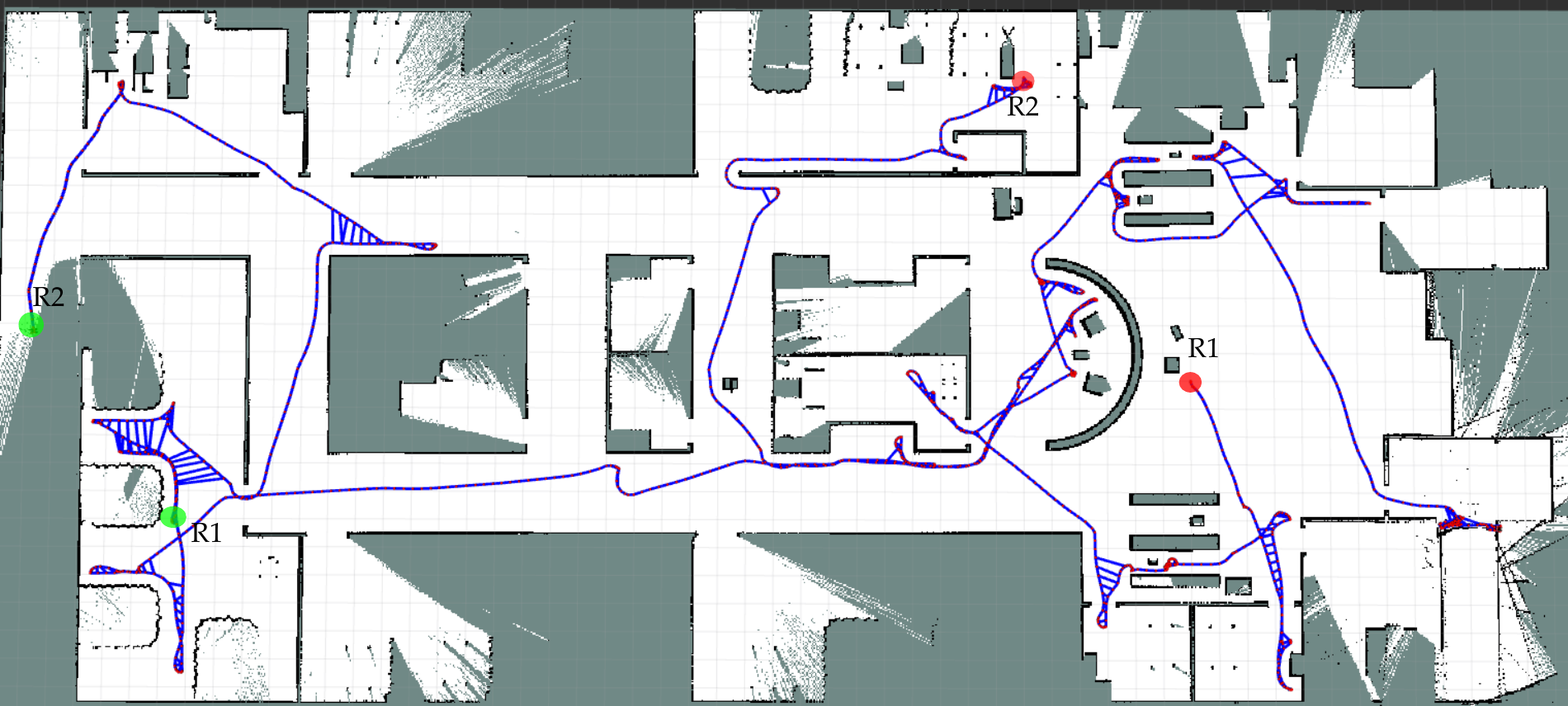}}      
            
      \caption{Environments used and the resulting O.G map showing the initial (red) and final positions (green) of robots.}
      \label{fig: plugin_map}
    \end{figure}

The simulations\footnote{YouTube link: https://youtu.be/MsZqoaEA0gY} were carried out on ROS Noetic, Gazebo, and Ubunto 20.04 on Intel Xeon\textsuperscript{\textregistered} W-2235 CPU 3.80GHz x 12, with 64Gb RAM and Nvidia Quadro RTX 4000 GPU. As described earlier we modified the approach of \cite{placed_fast_2021} to multi-robot and implemented the proposed approach as mentioned in Section \ref{sc: methodology} using Open Karto as SLAM backend, RosBot\footnote[2]{\url{https://husarion.com/}.} equipped with Lidar, and planners from the ROS navigation stack.
We used maps open-source maps of modified Willow Garage (W.G) from Gazebo simulator and AWS hospital environments (HOS)\footnote[3]{\url{https://github.com/aws-robotics}.} measuring 2072$m^2$ and 1243$m^2$ respectively. Figure \ref{fig: plugin_map} shows the Gazebo images and resulting O.G map of HOS indicating the initial, and final poses and resulting pose graphs. The ground truth O.G maps were generated using the gazebo\_2Dmap\_plugin\footnote[4]{\url {https://github.com/marinaKollmitz/gazebo}.} which uses wavefront exploration.  

We compared our proposed approach against: 1) Frontier Detection based Exploration (Frontier)\footnote[5]{\url{http://wiki.ros.org/frontier_exploration}.} of \cite{yamauchi_frontier-based_1997} which uses a greedy frontier exploration strategy without any SLAM uncertainty quantification. 2) and \cite{placed_fast_2021} by converting it into a multi-robot system namely MAGS. For environment exploration we debate on metrics of percentage of area covered, goal points reduction, percentage of unknown cells (\texttt{PER\_UNK}), and radius values (\texttt{RAD}). Regarding map quality, we compared metrics measuring Structural Similarity Index Measurement (SSIM) $\in [0,1]$, Root Mean Square Error (RMSE), and Alignment Error (AE) with reference to ground truth maps. We conducted 15 simulations of 20 minutes each for both W.G. and HOS using Frontier, MAGS, and our methods rendering a total simulation time of 15 hours. \texttt{PER\_UNK}, \texttt{RAD}, \texttt{MIN\_PTS} and \texttt{MAX\_PTS} were initialized to 60 \%, 1$m$, 0 and 10 respectively.

\begin{figure}
    \centering
      \subfloat[W.G\label{fig:map_all:1a}]{%
           \includegraphics[height=4cm ,width=4.31cm]{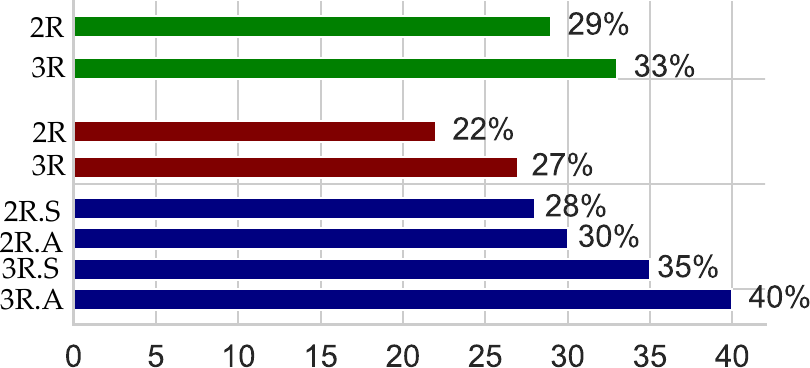}}
        \hfill
      \subfloat[HOS \label{fig:map_all:1b}]{%
            \includegraphics[height=4cm ,width=4.31cm]{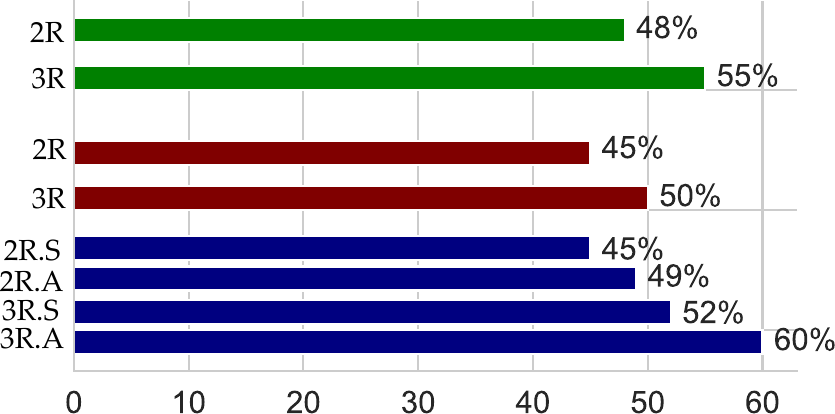}}  
            
      \caption{\% of map discovered in W.G and HOS Environments. }
      \label{fig: map_all}
    \end{figure}

\begin{figure}
    \centering
    \includegraphics[height=4cm ,width=8cm]{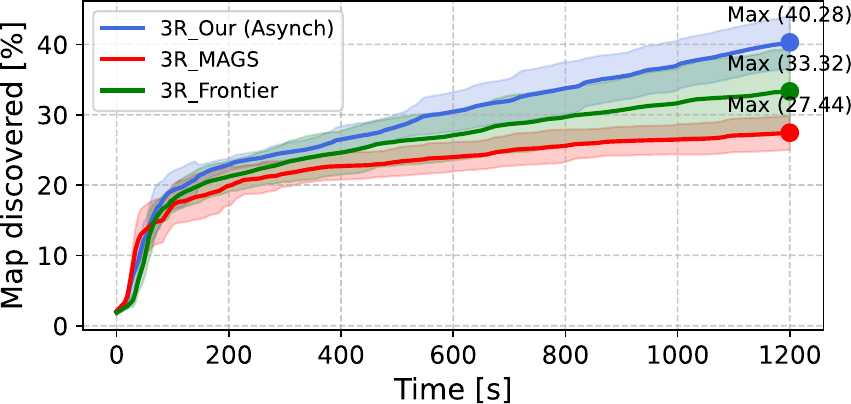}
    \caption{\% of map explored using Our, MAGS and Frontier approaches using 3 robots on W.G environment.}
    \label{fig: map_merged_asynch_2R3R}
\end{figure}

Figure \ref{fig: map_all} shows the percentage of maps discovered using Our approach (blue), MAGS (red), and Frontier (green) where R is the number of robots and S \& A denote synchronous and asynchronous approaches. It can be observed that Our approach covers an average of 10\%  and 7.5\% more area in W.G and HOS compared to MAGS and Frontier approaches and the agents controlled asynchronously were able to discover a higher percentage of the map concerning the ones controlled synchronously. Furthermore, we observe that Frontier outperforms MAGS as it performs frontier exploration without any uncertainty quantification, resulting in more exploration.  
Figure \ref{fig: map_merged_asynch_2R3R} shows the average rate of exploration for W.G along with the standard deviation using 3 robots. We can conclude that Our approach manages to explore 6.7\% and 13\% percentage more area than Frontier and MAGS.

\begin{figure}
    \centering
      \subfloat[W.G\label{fig: boxplots:1a}]{%
           \includegraphics[height=3.5cm ,width=4.31cm]{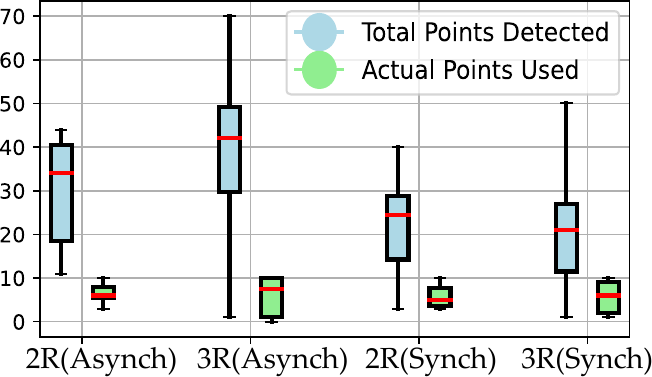}}
        \hfill
      \subfloat[HOS \label{fig: boxplots:1b}]{%
            \includegraphics[height=3.5cm ,width=4.31cm]{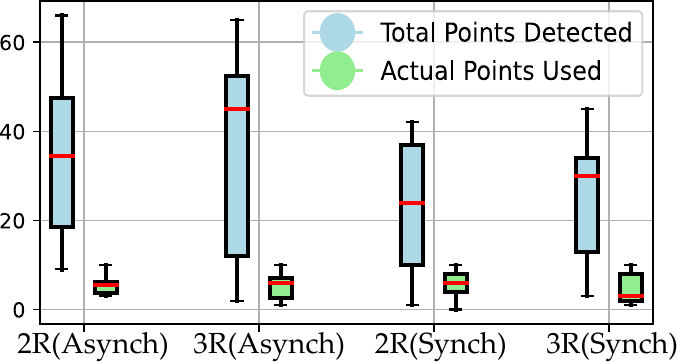}} 
           
      \caption{Number of Points vs the applied approach.}
      \label{fig: boxplots}
    \end{figure}

In Figure \ref{fig: boxplots} we can deduce that in both W.G and HOS the synchronous and asynchronous approach the number of points is reduced significantly, with a reduction of 80\%, 78\%, 80\%, and 65\% for all the cases respectively in W.G. And that of  85\%, 84\%, 72\% and 83\% for HOS. Consequently reducing the computational cost required by the reward processing on the server side with the adoption of frontier management strategies in Section \ref{subsec: frontiers_management} to limit the number of global frontiers. Furthermore, the average detected points are lower in synch as the agents have to wait for others, consequently lowering the overall number of points but with the cost of less exploration as evident from Figure \ref{fig: map_all}.

Table \ref{tab: PER_UNK} and Table \ref{tab: radius} shows the usage of  \texttt{PER\_UNK} i,e the percentage of unknown cells to be considered in the radius for computing the information gain of a frontier candidate and \texttt{RAD} showing the radius values changed when the list of points is recomputed using 3 robots with the async approach. We can observe that in W.G,  \texttt{PER\_UNK} remains at $\leq$ 40\% indicating the re-computation of the list because W.G has fewer obstacles than HOS resulting in less frontier neighbour percentage. Furthermore, we observe that the percentage of \texttt{RAD} and  \texttt{PER\_UNK} remain at 1.00 and $\leq$ 40\% respectively indicating less re-computation of the list on the server, consequently lowering the computational cost.  

Regarding the visual analysis of the maps and debating on map quality matrices, the results on average appear promising as shown in (Table \ref{tb: accuracy_comparison}) using 3 robots. In almost all cases, using Our method rendered reduced RMSE, AE, and increased SSIM as compared to MAGS and Frontier methods. We further conclude that the Frontier method when compared to MAGS explores the environment more as shown in Figure \ref{fig: map_all}, but has higher RMSE, AE, and lower SSIM.

\begin{table}
\begin{minipage}[b]{42mm}
    \caption{\texttt{PER\_UNK} usage} 
     \label{tab: PER_UNK}
\begin{tabular}{|c|c|c|}
        \hline
        \textbf{Env.} &  \textbf{\texttt{PER\_UNK}} & \textbf{Used} \\
        \hline        
        \multirow{3}{*}{W.G}
              &   60 \%  & 34.5 \%   \\	
               &   50 \%  & 1.4 \%    \\	
               &   $\leq$ 40 \%  & \textbf{64.0} \%   \\	             
        \hline
        \multirow{3}{*}{HOS}
              &   60 \%  & \textbf{67.3} \%    \\	
               &   50 \%  & 4.3 \%    \\	
               &   $\leq$ 40 \%  & 28.2 \%   \\	
        \hline               
\end{tabular}		
        \label{tb: accuracy_comparison}
\end{minipage}
\begin{minipage}[b]{42mm}
    \caption{\texttt{RAD} usage} 
     \label{tab: radius}
\begin{tabular}{|c|c|c|}
        \hline
        \textbf{Env} &  \textbf{\texttt{RAD}} & \textbf{Used} \\
        \hline        
        \multirow{3}{*}{W.G}
              &   1.00 & \textbf{87.0} \%   \\	
               &   1.25 & 1.8\%   \\	
               &   $\geq$ 1.50 & 9.7\%   \\	             
        \hline
        \multirow{3}{*}{HOS}
              &   1.00 & \textbf{76.2} \%   \\	
               &   1.25 & 5.1 \%   \\	
               &   $\geq$ 1.50 & 8.5 \%   \\	
        \hline               
		\end{tabular}		
\end{minipage}
\end{table}

\begin{table}
		\centering
  	\caption{MAP QUALITY METRICES.}
    
		\begin{tabular}{|c|c|c|c|c|}
        \hline
        \textbf{Env} &  \textbf{Method} & \textbf{SSIM} & \textbf{RMSE} &\textbf{AE} \\
        \hline        
        \multirow{1}{*}{W.G}
              &   Our (Asynch) & 0.74  & 5.43 & 25.68\\	
        \hline
        \multirow{1}{*}{W.G}
              & MAGS & \textbf{0.86}  & 6.34 & 28.39\\		
        \hline
        \multirow{1}{*}{W.G}
          & Frontier & \textit{0.20}  & \textbf{10.04} & \textbf{40.89}\\		
        \hline
        \multirow{1}{*}{HOS}
             & Our (Asynch) & \textbf{0.74}  & 4.89 & 25.39\\	        	
        \hline
        \multirow{1}{*}{HOS}
              & MAGS & 0.72  & 6.39 & 29.98\\	
         \hline
        \multirow{1}{*}{HOS}
              & Frontier & 0.35  & \textbf{12.67} & \textbf{42.89}\\	
         \hline
		\end{tabular}		
        \label{tb: accuracy_comparison}
	\end{table}

\subsection{Experimental Results}
\label{sc: experimental_results}
\begin{figure}
    \centering
      \subfloat[RosBot 2\label{fig10:1a}]{%
           \includegraphics[height=1.5cm ,width=2cm]{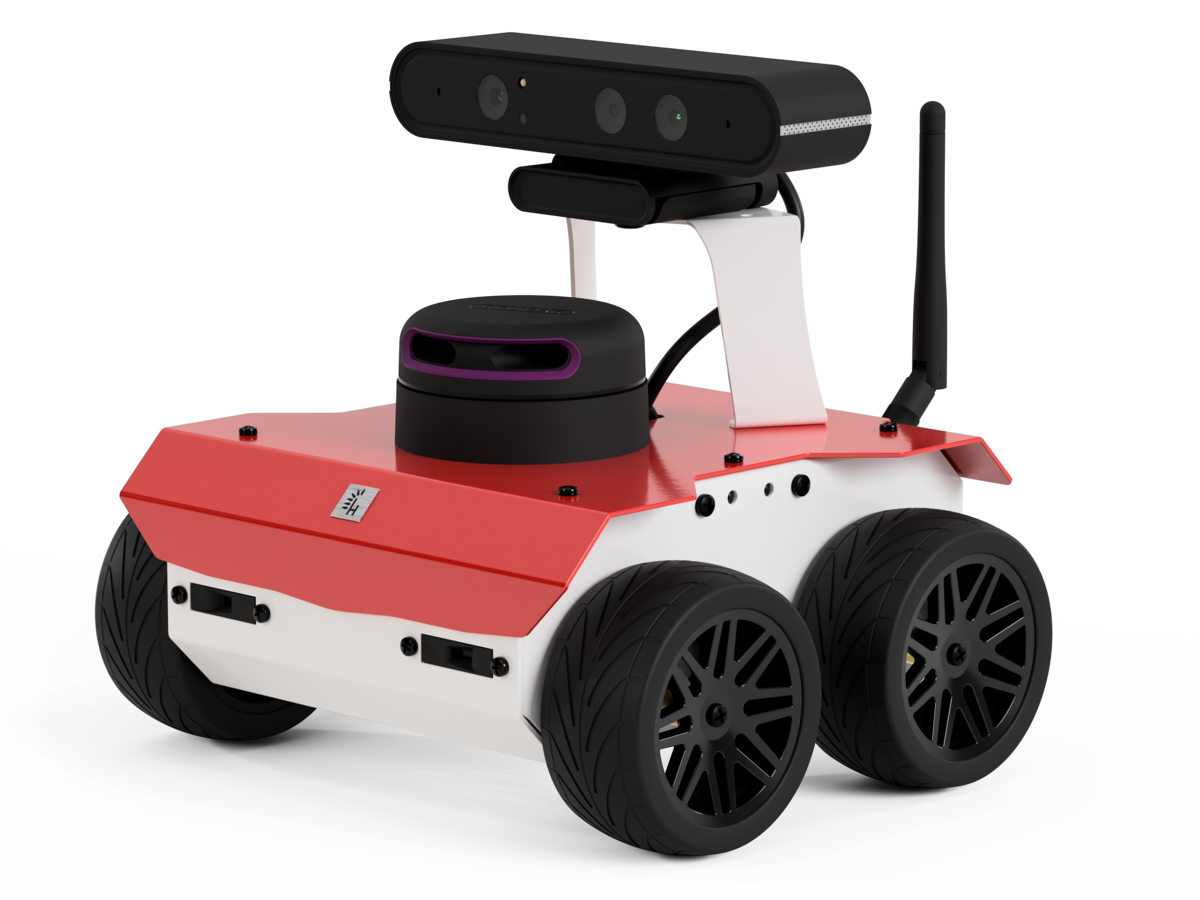}}
        \hfill
      \subfloat[Environment\label{fig10:1b}]{%
            \includegraphics[height=3cm ,width=6cm]{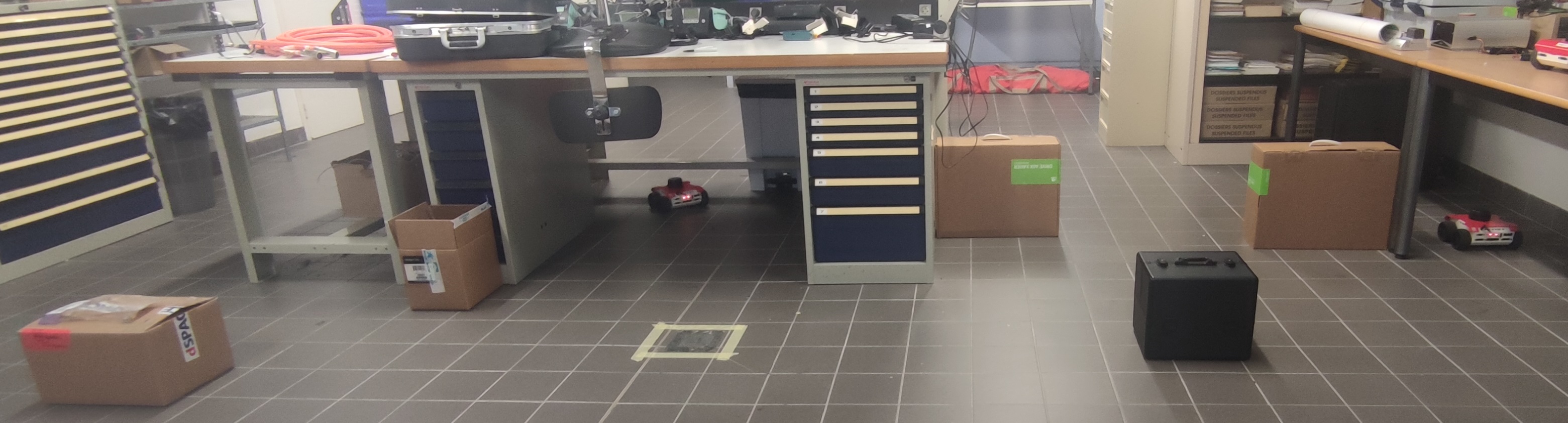}}                   
      \caption{Robot and experimental environment used.}
      \label{fig:10} 
    \end{figure}

\begin{figure}
    \centering
      \subfloat[Our\label{Rviz_final_map:1a}]{%
\includegraphics[height=3.5cm,width=5.1cm]{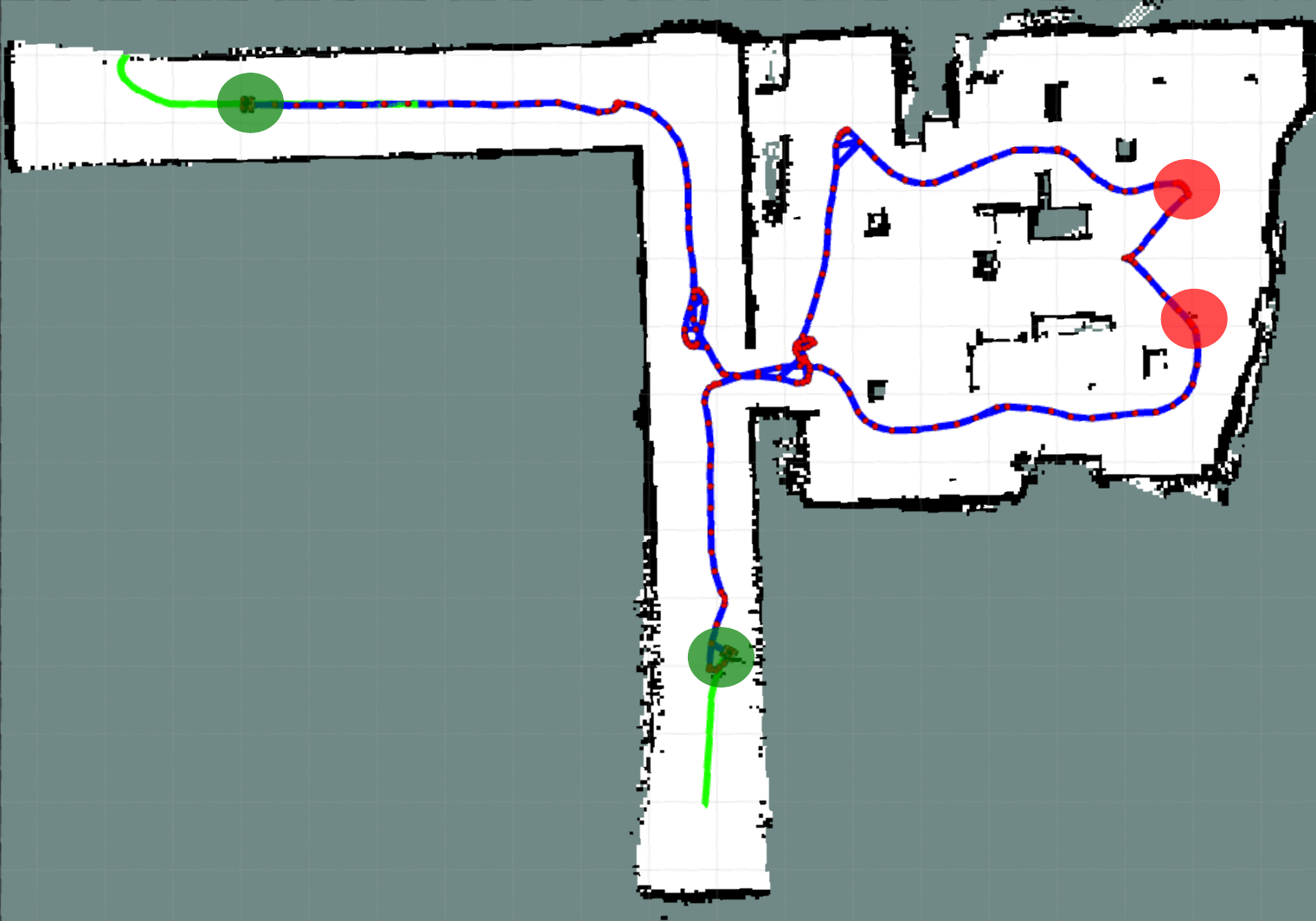}}
        \hfill
      \subfloat[MAGS\label{Rviz_final_map:1b}]{%
\includegraphics[height=3.5cm,width=3.5cm]{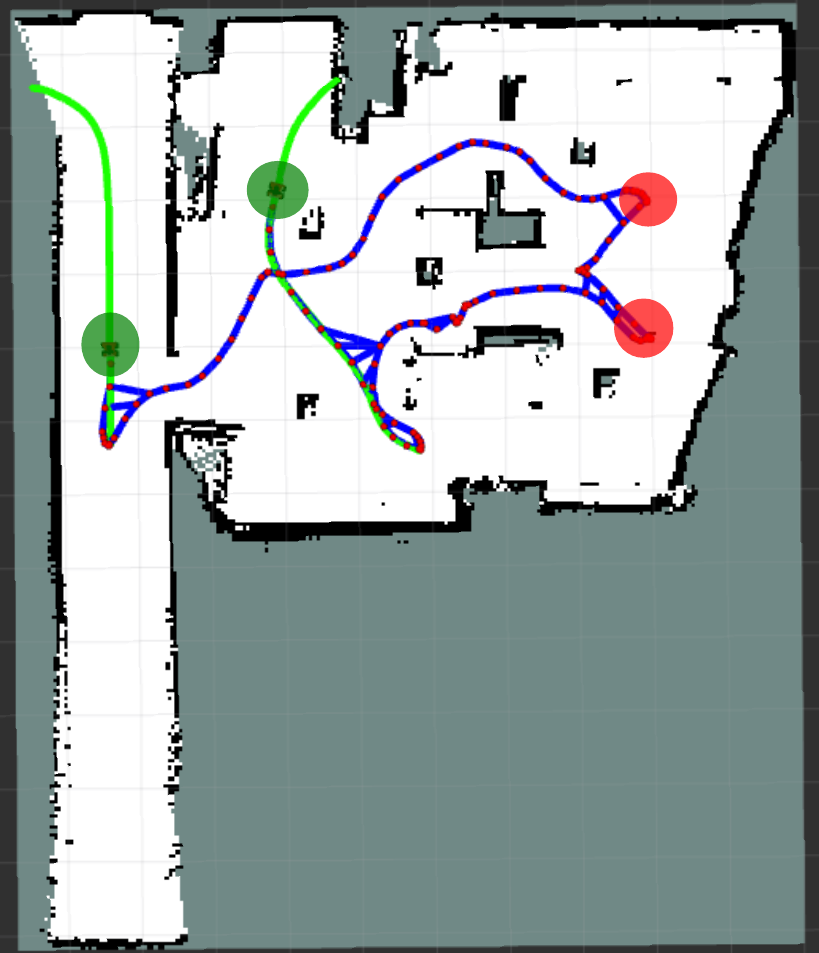}}                
      \caption{Final O.G map using Our and MAGS methods indicating initial (red) and final positions of agents (green)}
      \label{fig: Rviz_final_map} 
    \end{figure}

Experiments in a real environment were performed using two ROSBot 2R robots\footnote[5]{\url{https://husarion.com/manuals/rosbot/}.} with RPLidar A2 (Figure \ref{fig10:1a}) with ROS on Ubuntu 20.04.6 (LTS). The robots are equipped with an Intel Core i7\textsuperscript{\textregistered} CPU, with a system RAM of 32GB and NVIDIA RTX 1000 GPU. The environment consists of a room and two corridors measuring 81$m^2$ in total as shown in Figure \ref{fig10:1b}. Figure \ref{fig: Rviz_final_map} shows the resultant O.G map along with karto SLAM pose graphs using two robots. From Figure \ref{Rviz_final_map:1a} we can observe that using Our approach each agent effectively spreads and explores the environment with a total explored area of 70.31$m^2$ compared to that of only 55.80 $m^2$ for MAGS from Figure \ref{Rviz_final_map:1b}. We performed four experiments two using Our and two with MAGS with an experimental time of 20 minutes for each.

\begin{figure}
    \centering
    \includegraphics[height=3.5cm,width=8.5cm]{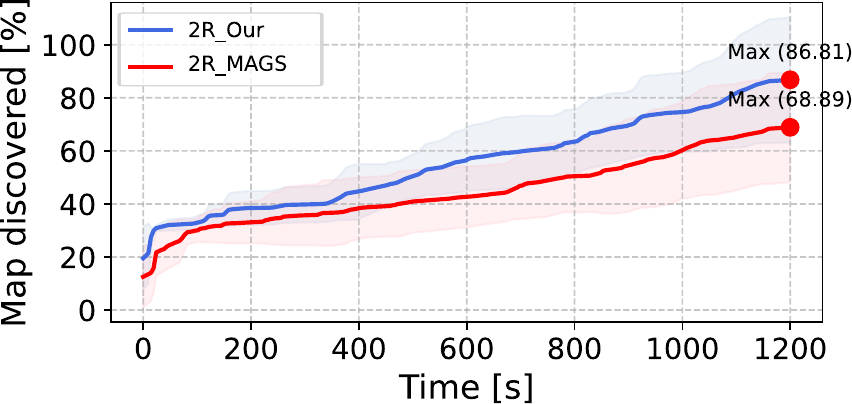}
    \caption{\% of explored map evolution in experiments}
    \label{fig: Figure_expresults}
\end{figure}

Figure \ref{fig: Figure_expresults} shows the average rate of percentage of the global map discovered by the robots with Our (2 experiments) and MAGS (2 experiments) methods respectively. It is evident that using Our approach we manage to cover 26\% more map area than MAGS.

\begin{figure}
    \centering
    \includegraphics[height=3cm,width=6cm]{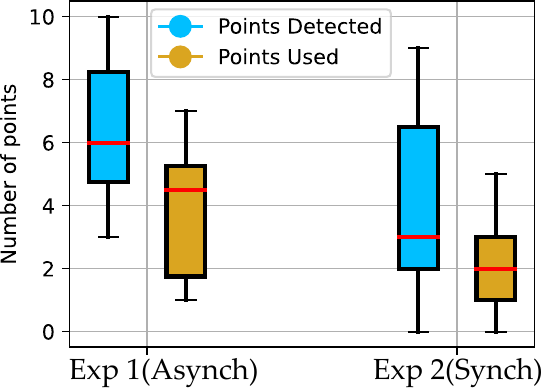}
    \caption{Points reduction in experiments}
    \label{fig: expBox_1}
\end{figure}

The box plot graph shown in Figure \ref{fig: expBox_1} shows a reduction in the number of processed points used for exploration using the asynchronous and synchronous methods using Our approach. We can observe that the average number of points reduces both methods from 6 to 5 and from 3 to 2 points respectively. We also observe that the number of points detected in the synchronous approach is less because the robots wait for each other to reach the goal before processing new points. Thus we observe more points in the asynchronous approach. We observe very less points in total as compared to the simulated environment because of the large difference in environment size. 

\section{CONCLUSIONS}
\label{sc: conclusions}
We proposed a method for the coordination of multiple robots in a collaborative exploration domain performing AC-SLAM. We proposed a strategy to efficiently manage the global frontiers to reduce the computational cost and to spread the robots in the environment. Two different coordinating approaches were presented for efficient exploration of the environment. We presented extensive simulation analysis on publicly available datasets and compared our approach to similar methods using ROS and performed experiments to validate the efficiency and usefulness of our approach in the real-world scenario. Possible future works can explore strategies to implement the proposed architecture in a decentralized way, thus dividing the computational weight among all the agents and using visual sensors for extracting features as potential frontier candidates. 

\addtolength{\textheight}{-8.5cm}   

\section*{ACKNOWLEDGMENT}
This work was conducted within the framework of the NExT Senior Talent Chair DeepCoSLAM, funded by the French Government through the program "Investments for the Future" administered by the National Agency for Research (ANR-16-IDEX-0007). We also extend our gratitude to the Région Pays de la Loire and Nantes Métropole for their invaluable support in facilitating this research endeavour.



\bibliographystyle{IEEEtran}

\bibliography{main}

\end{document}